\documentclass{article}

\usepackage[preprint]{neurips_2026}

\usepackage[utf8]{inputenc} 
\usepackage[T1]{fontenc}    
\usepackage{hyperref}       
\usepackage{url}            
\usepackage{booktabs}       
\usepackage{amsfonts}       
\usepackage{nicefrac}       
\usepackage{microtype}      
\usepackage{amsmath}
\usepackage[table]{xcolor}
\usepackage{graphicx}
\usepackage{wrapfig}
\usepackage[most]{tcolorbox}
\usepackage{listings}

\title{ProSR: Process-Shaped Spatial Reasoning for Reliable Chain-of-Thought in VLMs}

\author{Jiangyang Li$^{1}$ \quad
 Cong Wan$^{1}$ \quad
 Changjie Wu$^{2}$ \quad
 Songlin Dong$^{4}$\thanks{Corresponding author: Songlin Dong (dongsl@suat-sz.edu.cn)} \quad
 Lingjun Zhang$^{3}$ \\[0.1em]
 \textbf{Linzhe Shi$^{2}$} \enspace 
 \textbf{Xu Wang$^{2}$} \enspace 
 \textbf{Zhiheng Ma$^{4}$} \enspace
 \textbf{Hang Zhang$^{2}$} \enspace
 \textbf{Mu Xu$^{2}$} \enspace
 \textbf{Yihong Gong$^{1}$} \\[0.1em]
$^1$Xi’an Jiaotong University \enspace
$^2$Amap, Alibaba Group \enspace
$^3$Tsinghua University \\[0.1em]
$^4$Shenzhen University of Advanced Technology
\\ 
}

\begin{document}

\maketitle
\vspace{-10pt}

 \begin{abstract}

  Reliable spatial reasoning remains a core bottleneck for vision-language models (VLMs). Existing mainstream training paradigms for spatial reasoning largely rely on outcome alignment or process imitation, lacking explicit constraints on the reasoning process, and therefore struggle to ensure genuine visual dependence and stable reasoning trajectories. In this paper, we construct a high-quality CoT dataset covering diverse spatial phenomena and diagnose the model's reasoning process, revealing two typical types of process degradation during reinforcement learning optimization: \emph{Spurious Grounding}, which bypasses visual evidence, and \emph{Tail Instability}, where uncertainty abnormally rises in the later stage of reasoning. To address these issues, we propose \textbf{ProSR}, a process-shaping optimization framework for spatial reasoning. Through a \emph{Counterfactual Invariance Penalty} and a \emph{Tail Drift Penalty}, ProSR extends the optimization objective from single answer correctness to two process-level dimensions: visual dependence and trajectory stability. Experiments on multiple complex and out-of-distribution spatial reasoning benchmarks show that ProSR improves answer accuracy while generating reasoning trajectories that are more stable and more dependent on visual evidence.
  \end{abstract} 

 \section{Introduction}
\label{sec:intro}

VLMs have made significant progress on tasks such as general visual understanding, visual question answering, and cross-modal dialogue~\cite{alayrac2022flamingo,li2023blip,dai2023instructblip}. However, reliable spatial reasoning remains a long-standing weakness~\cite{liu2023visual,zhang2024vision,jia2025omnispatial}. Unlike general visual perception tasks, spatial reasoning requires models not only to recognize objects in images, but also to accurately understand relative positions, directions, occlusion, viewpoint changes, and multi-step compositional relationships~\cite{wan2026remot}. Such capabilities are crucial for frontier applications such as embodied intelligence, navigation~\cite{li2026trajectory}, and visual instruction following~\cite{zitkovich2023rt,kim2024openvla}. Therefore, improving the reliability of spatial reasoning in VLMs is of significant research value.

To truly improve spatial reasoning ability, the key is not only to make models ``answer correctly,'' but also to enable them to form reasoning processes that are stable and genuinely grounded in visual evidence. Existing mainstream methods typically improve performance by constructing large-scale spatial reasoning data with broad coverage~\cite{cai2025scaling}. However, these methods usually take final-answer correctness as the main optimization objective and lack explicit constraints on the reasoning process itself~\cite{ross2017right}. In contrast, CoT provides fine-grained intermediate steps such as object localization, relation comparison, and multi-step logical deduction, thereby offering a natural interface for explicitly modeling spatial reasoning processes~\cite{wei2022chain,zhang2023multimodal,shao2024visual}. Nevertheless, most existing CoT-related methods treat CoT merely as a static supervision signal, without further constraining its visual grounding ability and process stability~\cite{wu2025grounded,lanham2023measuring,turpin2023language}. In other words, limited by training paradigms based on ``outcome alignment'' or ``process imitation'', a model may generate seemingly plausible CoT even without truly relying on visual evidence, and its reasoning trajectory may also lack stability, thereby limiting its generalization ability in complex scenarios and out-of-distribution (OOD) tasks~\cite{agrawal2018don,dancette2021beyond,kv2020reducing}.

\begin{figure}[t]
\centering
\includegraphics[width=\linewidth]{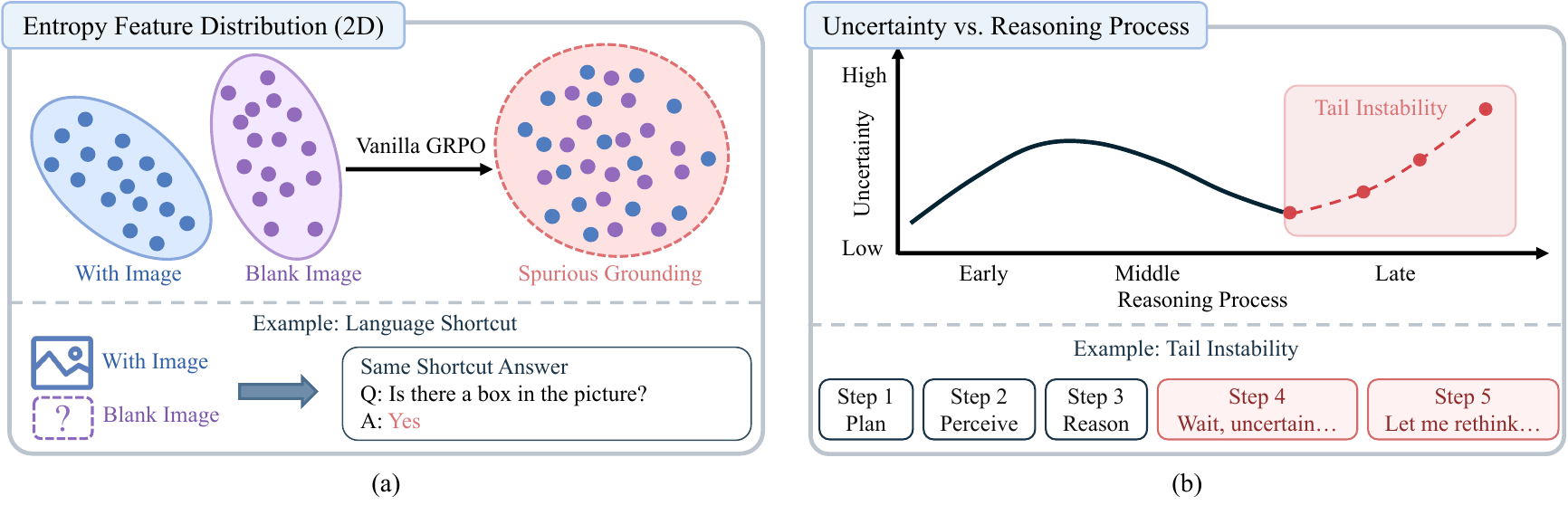}
\caption{
Process degradation in vanilla GRPO.
(a) Spurious Grounding: original- and blank-image rollouts become indistinguishable in entropy-feature space and produce the same answer despite missing visual evidence.
(b) Tail Instability: late-stage uncertainty re-rise exposes unstable reasoning trajectories.
}
\vspace{-15pt}
\label{fig:intro_diagnosis}
\end{figure}

To systematically study the learning process of spatial CoT reasoning and its potential degradation, we construct a CoT dataset covering multiple types of spatial reasoning scenarios and use this dataset to obtain a model with initial CoT capability through supervised fine-tuning (SFT). On this basis, we focus on the reinforcement learning stage: we adopt vanilla GRPO to further optimize the model, using only final-answer correctness as the reward signal~\cite{ouyang2022training,shao2024deepseekmath,guo2025deepseek}. As shown in Fig.~\ref{fig:intro_diagnosis}, we conduct a systematic analysis of the model's rollout behavior and find that this optimization process leads to two typical types of process degradation. The first is \emph{Spurious Grounding}: after visual information is removed, the model may still produce reasoning trajectories similar to those under the original image condition, indicating that the model may, to some extent, rely on language priors or dataset biases rather than sufficiently using visual evidence~\cite{li2023evaluating,guan2024hallusionbench}. The second is \emph{Tail Instability}: under the normal image condition, the model's uncertainty sometimes does not decrease monotonically as reasoning progresses, but instead rises again in the later part of the reasoning chain, indicating that the model may still exhibit certain instability when approaching answer generation~\cite{lanham2023measuring}. These phenomena suggest that simply pursuing final-answer correctness is insufficient to drive models to learn reliable spatial reasoning processes.

Based on these observations, we propose \textbf{ProSR} (\textbf{Pro}cess-Shaped \textbf{S}patial \textbf{R}easoning), a process-shaping optimization framework for spatial reasoning. Its core idea is to transform process defects in spatial CoT reasoning into optimizable reward constraints. For Spurious Grounding, we design a \emph{Counterfactual Invariance Penalty}: during the GRPO rollout process, we introduce a blank-image channel and penalize the similarity between the normalized entropy trajectories under the real-image and blank-image conditions, thereby reducing the model's reliance on non-visual shortcuts and guiding it to use visual evidence more. For Tail Instability, we introduce a \emph{Tail Drift Penalty}, which constrains abnormal increases in uncertainty during the later stage of reasoning and promotes stable convergence of the reasoning trajectory when approaching answer generation. Overall, we extend the optimization objective from single final-answer correctness to process constraints at two levels: visual dependence and trajectory stability, thereby encouraging the model to learn more robust and more generalizable spatial reasoning strategies.

We evaluate ProSR on multiple complex and out-of-distribution spatial reasoning benchmarks, achieving an average accuracy improvement of 3.7\% over the SOTA. Beyond answer-level performance, we further introduce four diagnostic metrics, namely Blank-image Accuracy, Same-Answer Rate, Normalized Trajectory Similarity, and Late-Rise Rate, to quantify Spurious Grounding and Tail Instability. The diagnostic results show that ProSR effectively mitigates these process-level degradation patterns, leading to stronger visual dependence and more stable reasoning trajectories.

Our contributions are as follows:
(1) We construct a CoT dataset covering multiple types of spatial reasoning scenarios and propose a reasoning-trajectory diagnostic protocol to characterize behavioral changes during model optimization. This protocol reveals two typical types of process degradation, Spurious Grounding and Tail Instability, indicating that relying solely on answer correctness is insufficient to guarantee reliable spatial reasoning.
(2) We propose ProSR, a process-shaping optimization framework for spatial reasoning. It formalizes the two degradation patterns as optimizable process constraints, and models counterfactual visual dependence and tail-stage reasoning stability through the Counterfactual Invariance Penalty and Tail Drift Penalty, respectively, thereby extending the learning objective from single answer correctness to process-level reliability.
(3) We validate the effectiveness of ProSR on multiple spatial reasoning benchmarks and diagnostic metrics. Experimental results show that ProSR not only improves answer accuracy, but also produces reasoning trajectories that are more stable and more strongly grounded in visual evidence.
 \section{Related Work}
  \label{sec:related}

  \textbf{Spatial Reasoning in Vision-Language Models.}
  Spatial reasoning~\cite{zheng2025learning,hu2025g,wu2025spatial} has long been recognized as a core challenge for vision-language models, particularly in tasks
  involving relative position understanding, viewpoint transformation, multi-view correspondence, and camera or object motion reasoning. Recent work has
  introduced a broad range of benchmarks~\cite{zhang2025flatland,du2024embspatial,wang2025site,yang2025thinking} and training resources~\cite{fan2025vlm,liu2026openspatial,cai2025scaling,chen2024spatialvlm} for evaluating and improving these abilities, spanning qualitative spatial
  relations, metric reasoning, and multi-view scene understanding. Some datasets emphasize 3D-aware or metric
  reasoning~\cite{tong2024cambrian,fu2024blink,ma20253dsrbench}, while others focus on cross-view alignment, egocentric motion, or object localization under
  viewpoint change~\cite{yang2025mmsi,yin2025spatial,li2025viewspatial}. While these efforts have substantially expanded the coverage of spatial reasoning tasks,
  they are still primarily centered on task performance, with relatively limited attention to the reliability of the underlying reasoning process.

  \textbf{Chain-of-Thought for Language and Vision-Language Reasoning.}
  Chain-of-thought prompting has shown that explicit intermediate reasoning can improve the performance of large language models, and has been extended to
  multimodal settings by prompting models to reason step by step over images and text~\cite{wan2026remot,chen2025think,li2026thinking,yang2025visual,cao2025spatialdreamer}. Related work has explored visual rationales that expose answer-
  relevant evidence~\cite{ouyang2025spacer,yin2025spatial}, as well as recent ``thinking'' vision-language models~\cite{bai2025qwen3} that produce longer
  reasoning traces. In spatial tasks, such rationales can help models identify visual anchors, compare relative relations, and compose multi-step transformations.
  However, existing CoT-based approaches are often used mainly to improve final accuracy, with less emphasis on whether the reasoning is concise, spatially
  grounded, and robust to visual perturbation. This gap motivates our focus on constructing spatially grounded CoT supervision and using it as a basis for
  process-level diagnosis.

  \textbf{Process Supervision and Reasoning Quality Evaluation.}
  Beyond outcome supervision, a growing line of work studies process supervision, verifier models, and step-level reasoning evaluation. Prior studies have
  analyzed uncertainty~\cite{paul2024making}, self-correction, entropy dynamics~\cite{li2026making,li2026entrocot,yan2026entrocut}, and reasoning trace
  quality~\cite{zhao2026entropy,jacovi2024chain,yu2025explainable} to better characterize model reasoning behavior. These results suggest that intermediate
  process signals can expose failure modes not captured by final-answer correctness alone. Nevertheless, most existing analyses focus on text-only domains such as
  mathematics, symbolic reasoning, or code, while process-level reliability in multimodal spatial reasoning remains less studied. Our work follows this process-
  oriented perspective, but extends it to spatial VLMs and further uses the resulting diagnostics to guide reward shaping in reinforcement learning.
 \section{Method}
  \label{sec:diagnosis}

\begin{figure}[t]
  \centering
  \includegraphics[width=\linewidth]{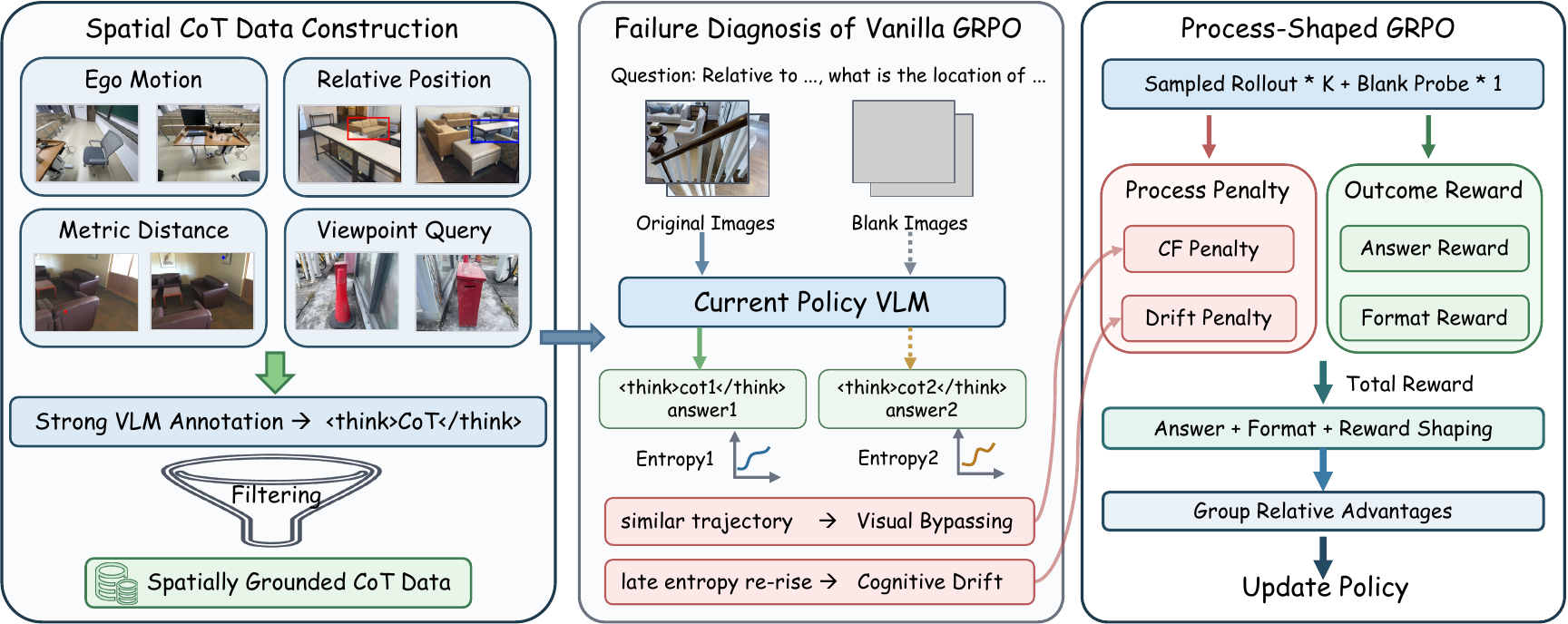}
  \caption{
    Overview of our diagnosis-driven framework for improving spatial reasoning in VLMs.
    We construct visually grounded CoT data, diagnose vanilla GRPO with paired original/blank-image rollouts and entropy-trajectory analysis, and translate the observed failures into process-level rewards.
    }
  \label{fig:pipeline}
  \end{figure}

  To improve the process-level reliability of VLMs on spatial reasoning tasks, we propose ProSR, a failure-diagnosis-driven reinforcement learning framework that turns reasoning-process deficiencies into reward signals. As shown in Figure~\ref{fig:pipeline}, we first construct concise and spatially grounded CoT supervision data, then diagnose
  the characteristic failure modes of vanilla GRPO, and finally translate these observations into process-level rewards that encourage visual dependence and
  trajectory stability. Section 3.1 describes the data construction procedure; Section 3.2 presents the failure diagnosis of vanilla GRPO; Section 3.3 introduces
  our diagnostic-guided reward shaping strategy; and Section 3.4 summarizes the training pipeline.

  \subsection{Spatial Reasoning Data Construction}

    We construct spatial reasoning CoT data by prompting Gemini-3.1-Pro-Preview to generate short, visually grounded rationales for spatial reasoning
  questions. Rather than eliciting verbose explanations, we ask the model to first identify the target spatial relation and then reason only over task-relevant
  visual evidence. Each CoT follows a standardized format, \texttt{<think></think><answer>X</answer>}, and is constrained to a small number of concise reasoning
  steps. This design encourages supervised fine-tuning to learn spatially anchored reasoning patterns instead of generic image descriptions or language-only
  explanations.

  After generation, we apply rule-based filtering to improve data quality. We keep only samples whose predicted answers match the ground truth, remove rationales
  with overly repetitive reconsideration or repeated sentences, and discard samples with insufficient spatial grounding. The exact filtering criteria, including
  the reconsider and spatial-anchor settings, are provided in Appendix~\ref{app:data}. Although these rules are imperfect, they effectively bias the training set
  toward concise rationales with higher spatial information density and stronger visual grounding.

  \subsection{Diagnosing Failure Modes in Vanilla GRPO}
  \label{sec:vanilla_diagnosis}

  Vanilla GRPO optimizes sampled rollouts mainly through outcome-level rewards, i.e., whether the final answer is correct and whether the output format is valid.
  Given a question \(x\), an image set \(I\), a reference answer \(y\), and a rollout \(r\sim\pi_\theta(\cdot\mid x,I)\), the basic reward is
  \begin{equation}
  R_{\mathrm{base}}(r;x,I,y)
  =
  R_{\mathrm{acc}}(r;y)
  +
  \lambda_{\mathrm{fmt}}R_{\mathrm{fmt}}(r),
  \end{equation}
  where \(R_{\mathrm{acc}}\) checks the final answer, \(R_{\mathrm{fmt}}\) enforces the target format, and \(\lambda_{\mathrm{fmt}}\) is the format weight. This
  objective improves answer quality, but does not directly reveal whether the model truly relies on visual evidence or whether its reasoning remains stable.

  To expose these hidden process-level failures, we build a balanced diagnostic set \(\mathcal{D}_{\mathrm{diag}}=\{(x_i,I_i,y_i)\}_{i=1}^{N}\) and perform paired
  rollout analysis. For each example, we sample
  \begin{equation}
  r_i \sim \pi_\theta(\cdot\mid x_i,I_i),
  \qquad
  \tilde r_i \sim \pi_\theta(\cdot\mid x_i,\tilde I_i),
  \end{equation}
   where \(\tilde I_i\) is obtained by replacing each image in \(I_i\) with a size-matched blank image. Let \(\hat y(r)\) denote the parsed final answer of rollout
  \(r\), and let \(\mathbf{e}^{r}=(e_1^r,\ldots,e_L^r)\) denote the token-level entropy trajectory within its thinking span. Concretely, the diagnostic subset
  contains 480 examples sampled from the same source pool as our training data and covers a diverse set of spatial task groups, so that the diagnostic statistics
  are not dominated by any single task type. Besides the standard image-conditioned accuracy \(A_{\mathrm{img}}\), we use four diagnostic metrics.

  \paragraph{(1) Blank-image Accuracy.}
  We measure how often the model remains correct after visual content is removed:
  \begin{equation}
  A_{\mathrm{blank}}
  =
  \frac{1}{N}\sum_{i=1}^{N}
  \mathbf{1}\!\left[\hat y(\tilde r_i)=y_i\right].
  \end{equation}
  A high \(A_{\mathrm{blank}}\) suggests that the model can still solve the task from language priors or dataset biases even without visual evidence.

  \paragraph{(2) Same-Answer Rate.}
  We further measure answer-level counterfactual invariance:
  \begin{equation}
  \mathrm{SAR}
  =
  \frac{1}{N}\sum_{i=1}^{N}
  \mathbf{1}\!\left[\hat y(r_i)=\hat y(\tilde r_i)\right].
  \end{equation}
  While \(A_{\mathrm{blank}}\) reflects task success under blank input, \(\mathrm{SAR}\) directly captures whether the model tends to preserve the same decision
  after the image is blanked out.

  \paragraph{(3) Normalized Trajectory Similarity.}
  Answer invariance alone does not tell whether the reasoning process also remains insensitive to the missing image. We therefore compare the \emph{shape} of the
  entropy trajectories at the sample level. Specifically, each trajectory is resampled to a fixed length \(T\) and \(\ell_1\)-normalized as \(\mathbf{z}^{r}
  =\mathcal{R}_{T}(\mathbf{e}^{r}) / (\|\mathcal{R}_{T}(\mathbf{e}^{r})\|_{1}+\epsilon)\). We then define
  \begin{equation}
  \mathrm{NTS}
  =
  \frac{1}{N}\sum_{i=1}^{N}
  \left(
  1-\frac{1}{2}\left\|
  \mathbf{z}^{r_i}
  -
  \mathbf{z}^{\tilde r_i}
  \right\|_1
  \right).
  \end{equation}
  Unlike raw entropy differences, \(\mathrm{NTS}\) ignores absolute scale and focuses on whether the temporal rhythm of uncertainty remains similar after visual
  evidence is removed.
  
   \paragraph{(4) Late-Rise Rate.}
  To diagnose late-stage instability, we divide the entropy trajectory of the image-conditioned rollout \(r_i\) into early, middle, and late segments, and define
  \begin{align}
  \Delta_{\mathrm{tail}}(r)
  &=
  \left[
  \mu_{\mathrm{late}}(\mathbf{e}^{r})
  -
  \mu_{\mathrm{mid}}(\mathbf{e}^{r})
  -
  m
  \right]_{+}, \\
  \mathrm{LRR}@\tau
  &=
  \frac{1}{N}\sum_{i=1}^{N}
  \mathbf{1}\!\left[\Delta_{\mathrm{tail}}(r_i)>\tau\right],
  \end{align}
  where \(m\) is a small margin and \([\,\cdot\,]_{+}\) denotes the positive part. A high \(\mathrm{LRR}@\tau\) indicates frequent late-stage entropy re-rise,
  suggesting unstable or ineffective reasoning near the end of generation.

  Together, \(A_{\mathrm{blank}}\), \(\mathrm{SAR}\), and \(\mathrm{NTS}\) diagnose spurious grounding from the levels of task outcome, final decision, and
  reasoning trajectory, respectively, while \(\mathrm{LRR}@\tau\) captures tail drift through late-stage entropy re-rise. These diagnostics make the failure
  modes of vanilla GRPO directly observable and directly motivate the two shaping terms introduced next: the counterfactual penalty for Spurious Grounding and the
  drift penalty for Tail Instability.

  \subsection{Diagnostic-guided Reward Shaping}
  \label{sec:diagnostic_reward}

  Beyond the answer and format rewards in Sec.~3.2, we introduce two entropy-trajectory-based shaping terms that directly target the two diagnosed failure modes
  of vanilla GRPO: the counterfactual invariance term targets Spurious Grounding, and the drift term targets Tail Instability. For a rollout \(r\), let \(\mathbf{e}^{r}=(e_1^r,\ldots,e_L^r)\) denote the token-level entropy sequence
  over its reasoning span. From \(\mathbf{e}^{r}\), we derive two complementary views: a normalized global trajectory for counterfactual comparison, and coarse
  stage-wise statistics for drift detection.

  \subsubsection{Counterfactual Invariance Penalty}
  \label{sec:cf_penalty}

  For each training question, besides the original-image rollout \(r\), we sample a matched blank-image rollout \(r^{\emptyset}\). Let \(\mathbf{z}^{r}\) and
  \(\mathbf{z}^{r^{\emptyset}}\) denote the resampled and normalized entropy trajectories defined above. Their sample-level normalized trajectory similarity is
  \begin{equation}
  s_{\mathrm{cf}}(r,r^{\emptyset})
  =
  1-\frac{1}{2}
  \left\|
  \mathbf{z}^{r}
  -
  \mathbf{z}^{r^{\emptyset}}
  \right\|_1
  \in [0,1].
  \end{equation}
  A large \(s_{\mathrm{cf}}\) means that the model follows nearly the same uncertainty rhythm even after visual evidence is removed. To avoid encouraging
  arbitrary divergence, we activate this term only when the original and blank-image rollouts produce the same final answer, and apply the bounded penalty
  \begin{equation}
  R_{\mathrm{cf}}
  =
  -
  \operatorname{clip}_{[0,1]}
  \left(
  \frac{s_{\mathrm{cf}}-\tau_{\mathrm{cf}}}{1-\tau_{\mathrm{cf}}}
  \right).
  \end{equation}
  Thus, only overly invariant trajectory shapes are suppressed, while moderate differences or genuine answer changes remain unconstrained.

  \subsubsection{Tail Drift Penalty}
  \label{sec:drift_penalty}

  To detect late-stage instability within a single rollout, we partition its reasoning span into early, middle, and late segments, and compute the corresponding
  mean entropies \(H_E^r\), \(H_M^r\), and \(H_L^r\), where \(H_s^r=\frac{1}{|\mathcal{I}_s|}\sum_{t\in\mathcal{I}_s} e_t^r\). We then penalize significant late-
  stage re-expansion of uncertainty:
  \begin{equation}
  R_{\mathrm{drift}}
  =
  -
  \operatorname{clip}_{[0,1]}
  \left(
  [H_L^r-H_M^r-m]_+
  \right).
  \end{equation}
  Here \(m\) is a tolerance margin used to ignore minor fluctuations. This term does not enforce globally monotonic entropy decay; it only discourages the
  specific pattern in which uncertainty rises again near the end and indicates unproductive tail-end search.

  \subsubsection{Final Optimization Objective}
  \label{sec:final_objective}

   The final reward is
  \begin{equation}
  R
  =
  R_{\mathrm{acc}}
  +
  \lambda_{\mathrm{fmt}}R_{\mathrm{fmt}}
  +
  \lambda_{\mathrm{cf}}R_{\mathrm{cf}}
  +
  \lambda_{\mathrm{drift}}R_{\mathrm{drift}},
  \end{equation}
  and GRPO maximizes the expected reward over training samples and sampled rollouts. In implementation, both shaping terms are evaluated only on valid reasoning
  spans, while exceptionally short or malformed outputs are handled by simple validity safeguards. Since \(R_{\mathrm{cf}}\) and \(R_{\mathrm{drift}}\) are
  bounded negative terms, they act as diagnostic constraints that complement, rather than replace, the answer-level optimization signal.

  \subsection{Training Pipeline}

   We adopt a two-stage training pipeline. In the first stage, we perform supervised fine-tuning on 22K filtered spatial reasoning CoT samples collected from
  MindCube~\cite{yin2025spatial}, SenseNova-800K~\cite{cai2025scaling}, and SPAR-7M~\cite{zhang2025flatland}. These samples are first annotated by Gemini-3.1-Pro-Preview using the same prompting protocol across the three source datasets,
  and then filtered into a spatially grounded SFT set. This stage teaches the model the structured reasoning-answer format and provides an initial spatially
  grounded reasoning prior, so that the policy does not enter reinforcement learning from a weak or unstructured initialization.

  In the second stage, we further optimize the model with GRPO on 45K spatial reasoning samples drawn from the same source pool, with the SFT subset expanded by
  additional training examples. The reward combines the conventional answer and format rewards with the diagnostic-guided process rewards introduced in
  Sec.~3.3. This design separates capability acquisition from process refinement: SFT provides a stable spatial
  reasoning initialization, while GRPO shapes the policy toward stronger visual dependence and greater trajectory stability.
\section{Experiments}
\label{sec:experiments}

\subsection{Experimental Setup}
\label{sec:exp_setup}

  \textbf{Implementation Details.}
  We use \textsc{Qwen3-VL-8B-Thinking}~\cite{bai2025qwen3} as the base model for all trainable variants. Following the two-stage pipeline in Sec.~3.4, we first perform SFT using
  AdamW, a global batch size of \texttt{8}, a learning rate of \texttt{5e-6}, and a maximum sequence length of \texttt{8192}. The resulting SFT checkpoint is used
  to initialize both vanilla GRPO and our ProSR. During reinforcement learning, we sample \texttt{8} rollouts for each prompt and optimize the policy
  with a learning rate of \texttt{2e-6}; a KL penalty with coefficient \texttt{0.04} is applied to constrain deviation from the reference model. Vanilla GRPO uses
  the standard combination of answer correctness and format rewards. Our method further incorporates the counterfactual invariance penalty and the tail drift
  penalty, weighted by \(\lambda_{\mathrm{cf}}=\texttt{0.1}\) and \(\lambda_{\mathrm{drift}}=\texttt{0.1}\), respectively. We set \(\lambda_{\mathrm{fmt}}
  =\texttt{0.2}\), \(\tau_{\mathrm{cf}}=\texttt{0.4}\), \(m=\texttt{0.1}\), and partition each thinking trajectory into early/middle/late segments with ratio
  \texttt{3:4:3}. Unless otherwise specified, all RL variants use the same training setup and evaluation protocol.

\textbf{Evaluation.}
We evaluate our method on five spatial reasoning benchmarks: \textsc{3DSRBench}~\cite{ma20253dsrbench}, \textsc{MindCube-Tiny}~\cite{yin2025spatial}, \textsc{ViewSpatial}~\cite{li2025viewspatial}, \textsc{EmbSpatial}, and \textsc{SPAR-Bench}~\cite{zhang2025flatland}. These benchmarks cover complementary spatial abilities, including multi-image, multi-view, egocentric, embodied, and fine-grained relation reasoning. We follow each benchmark's official protocol, report accuracy-based metrics, and use their average score as the overall indicator. For fair comparison, all methods use the same prompting, decoding, and answer extraction settings.

 \subsection{Main Results}
  \label{sec:main_results}

  \begin{table*}[t]
\centering
\small
\setlength{\tabcolsep}{6.2pt}
\renewcommand{\arraystretch}{1.15}
\caption{
Main results on five spatial reasoning benchmarks.
The best and second-best results among all non-reference models are highlighted
in \textbf{bold} and underlined, respectively.
}
\label{tab:main_results}
\vspace{4pt}
\resizebox{\textwidth}{!}{
\begin{tabular}{lcccccc}
\toprule
\rowcolor{gray!10}
\textbf{Model} &
\textbf{3DSRBench} &
\textbf{MindCube-Tiny} &
\textbf{ViewSpatial} &
\textbf{EmbSpatial} &
\textbf{SPAR-Bench} &
\textbf{Avg.} \\
\midrule

\multicolumn{7}{l}{\cellcolor{gray!8}\textit{Reference}} \\
Human
& 95.7 & 94.5 & -- & 90.3 & 67.3 & -- \\
Random Choice
& 20.9 & 33.0 & 26.3 & 25.0 & 32.7 & -- \\

\midrule
\multicolumn{7}{l}{\cellcolor{blue!6}\textit{Proprietary Models}} \\
Seed 1.6~\cite{guo2025seed1}
& 56.9 & 48.8 & 43.9 & 75.4 & 50.1 & 55.0 \\
Gemini 2.5 Pro~\cite{team2023gemini}
& 59.3 & 57.6 & 46.1 & 78.8 & 49.8 & 58.3 \\
GPT-5.2~\cite{singh2025openai}
& 60.2 & 60.4 & 47.3 & 79.3 & 55.1 & 60.5 \\
Gemini 3 Pro~\cite{team2023gemini}
& \textbf{68.9} & 70.9 & 50.4 & \textbf{84.3} & 48.7 & 64.6 \\

\midrule
\multicolumn{7}{l}{\cellcolor{green!6}\textit{Open-source General Models}} \\
Qwen2.5-VL-7B-Instruct~\cite{bai2025qwen2}
& 47.5 & 36.0 & 36.9 & 71.8 & 33.8 & 45.2 \\
BAGEL-7B-MoT~\cite{deng2025emerging}
& 50.2 & 34.7 & 41.3 & 73.1 & 39.1 & 47.7 \\
Qwen3-VL-8B-Instruct~\cite{bai2025qwen3}
& 53.9 & 29.4 & 42.2 & 77.7 & 39.6 & 48.6 \\
InternVL3\_5-8B~\cite{wang2025internvl3}
& 49.2 & 40.2 & 40.0 & 75.7 & 38.2 & 48.7 \\
Qwen3-VL-8B-Thinking~\cite{bai2025qwen3}
& 55.2 & 40.6 & 43.9 & 78.9 & 33.0 & 50.3 \\

\midrule
\multicolumn{7}{l}{\cellcolor{orange!8}\textit{Open-source Spatial Intelligence Models}} \\
SpatialLadder-3B~\cite{li2025spatialladder}
& 42.8 & 43.5 & 39.9 & 58.2 & 32.9 & 43.5 \\
SpaceR-SFT-7B~\cite{ouyang2025spacer}
& 40.5 & 38.0 & 35.9 & 66.9 & 34.2 & 43.1 \\
Cambrian-S-7B~\cite{yang2025cambrian}
& 45.0 & 37.9 & 41.3 & 72.8 & 37.9 & 47.0 \\
VST-7B-SFT~\cite{yang2025visual}
& 53.1 & 39.7 & 50.5 & 73.7 & 46.6 & 52.7 \\
SenseNova-SI-1.1-Qwen3-VL-8B~\cite{cai2025scaling}
& 53.2 & 73.8 & \underline{51.2} & 72.5 & 40.8 & 58.3 \\
GeoThinker~\cite{li2026thinking}
& 51.9 & \underline{83.0} & 45.9 & 78.8 & \textbf{68.2} & \underline{65.6} \\

\midrule
\multicolumn{7}{l}{\cellcolor{red!7}\textit{Ours}} \\
\textsc{Qwen3-VL-8B-Thinking} + SFT
& 57.3 & 68.4 & 45.6 & 78.6 & 54.3 & 60.8 \\
 \textsc{Qwen3-VL-8B-Thinking} + GRPO
  & 59.1 & 75.8 & 47.8 & 79.0 & 58.4 & 64.0 \\
\rowcolor{red!4}
\textbf{\textsc{Qwen3-VL-8B-Thinking} + Ours}
& \underline{62.4} & \textbf{88.6} & \textbf{51.4} & \underline{79.8} & \underline{64.3} & \textbf{69.3} \\

\bottomrule
\end{tabular}
}
\end{table*}
  \begin{figure}[t]
\centering
\includegraphics[width=\linewidth]{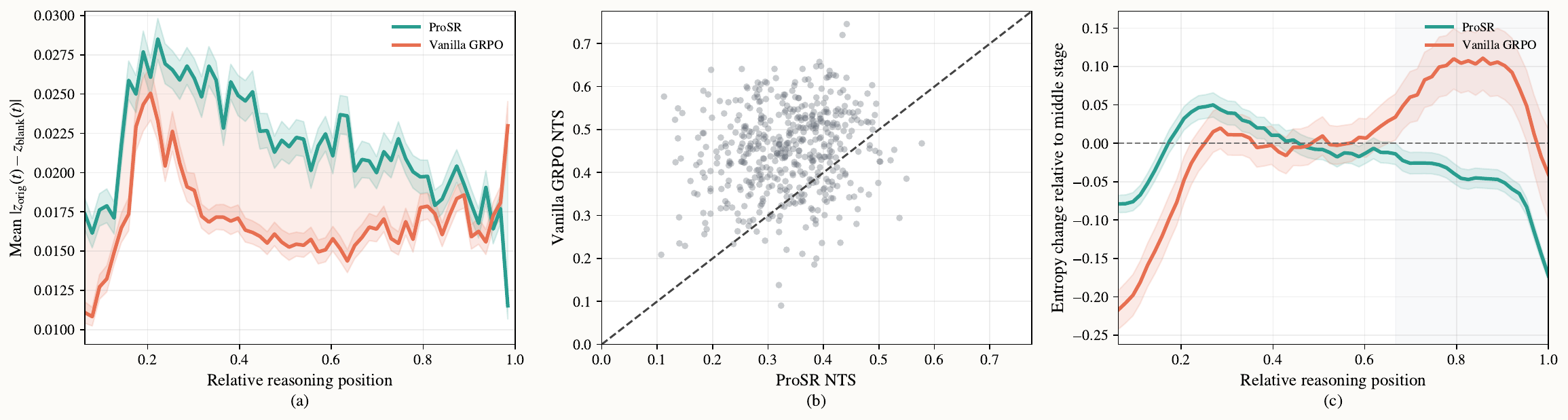}
\caption{
Process-level diagnostics on the balanced diagnostic subset.
(a) Counterfactual trajectory gap.
(b) Per-sample NTS of vanilla GRPO versus ProSR; points above the diagonal favor our method.
(c) Entropy change relative to the middle stage, highlighting late-stage uncertainty in original-image rollouts.
Shaded bands denote standard errors.
}
\label{fig:diagnostic_overview}
\end{figure}

  Table~\ref{tab:main_results} summarizes the performance of proprietary models, open-source generalist models, spatial intelligence models, and our variants on five spatial
  reasoning benchmarks. Among the trainable \textsc{Qwen3-VL-8B-Thinking} variants, ProSR consistently outperforms both the SFT-only model and vanilla GRPO, showing that answer-
  level reinforcement learning alone is insufficient and that process-level shaping brings additional gains.

  Our final model achieves the best overall performance, with an average score of 69.3 across the five benchmarks, outperforming the strongest baseline, GeoThinker, by 3.7 points.
  Compared with the SFT-only variant and vanilla GRPO, ProSR improves the average score from 60.8 to 69.3 and from 64.0 to 69.3, respectively, indicating that the gains come from
  both supervised spatial reasoning initialization and subsequent process-shaped reinforcement learning.

  The improvements are especially notable on \textsc{3DSRBench}, where ProSR reaches 62.4, surpassing all open-source baselines and ranking second overall behind Gemini 3 Pro. It
  also achieves 88.6 on \textsc{MindCube-Tiny}, the best score in the table, and 51.4 on \textsc{ViewSpatial}, again leading all evaluated models. On \textsc{SPAR-Bench}, ProSR
  remains strong with 64.3, ranking second only to GeoThinker. These results suggest that ProSR improves structured spatial reasoning and generalizes across diverse spatial
  benchmarks.

  Although closed-source models remain competitive on \textsc{3DSRBench} and \textsc{EmbSpatial}, ProSR achieves the highest average score overall, showing that targeted process-
  level optimization can narrow the gap with proprietary systems while surpassing existing open-source spatial reasoning models.

\subsection{Effect on Diagnostic Failure Modes}
\label{sec:diagnostic_effect}

We next examine whether the gains of ProSR are accompanied by improved process-level reliability. Using the balanced diagnostic subset from Sec.~3.2, we compare SFT, vanilla GRPO, and ProSR under the same paired original-image and blank-image rollout protocol.

\begin{table}[t]
  \centering
  \small
  \setlength{\tabcolsep}{5.5pt}
  \renewcommand{\arraystretch}{1.15}
  \caption{
  Effect on diagnostic failure modes measured on the balanced diagnostic subset.
  Lower values indicate fewer process-level failures.
  }
  \label{tab:diagnostic_effect}
  \vspace{3pt}
  \begin{tabular}{lcccc}
  \toprule
  \rowcolor{gray!10}
  \textbf{Method} &
  \(\boldsymbol{A_{\mathrm{blank}}}\downarrow\) &
  \(\boldsymbol{\mathrm{SAR}}\downarrow\) &
  \(\boldsymbol{\mathrm{NTS}}\downarrow\) &
  \(\boldsymbol{\mathrm{LRR}@0.1}\downarrow\) \\
  \midrule
  SFT
  & 0.3688 & 0.3854 & 0.3349 & 0.0042 \\
  Vanilla GRPO
  & 0.5354 & 0.5792 & 0.4516 & 0.1792 \\
  \rowcolor{red!4}
  \textbf{ProSR}
  & \textbf{0.3214} & \textbf{0.3379} & \textbf{0.2916} & \textbf{0.0017} \\
  \bottomrule
  \end{tabular}
\end{table}

 \begin{table}[t]
    \centering
    \small
    \setlength{\tabcolsep}{5.2pt}
    \renewcommand{\arraystretch}{1.15}
    \caption{
    Failure-aware breakdown on the diagnostic subset.
    Samples are grouped by vanilla GRPO failure severity under the paired original/blank protocol.
    Higher is better.
    }
    \label{tab:failure_breakdown}
    \vspace{3pt}
    \begin{tabular}{lcccc}
    \toprule
    \rowcolor{gray!10}
    \textbf{Failure group} &
    \(\boldsymbol{n}\) &
    \(\boldsymbol{\mathrm{Vanilla\ GRPO}}\uparrow\) &
    \(\boldsymbol{\mathrm{ProSR}}\uparrow\) &
    \(\boldsymbol{\Delta}\uparrow\) \\
    \midrule
    Clean
    & 235 & 89.4 & 90.2 & +0.8 \\
    \rowcolor{gray!4}
    Spurious-Grounding-only
    & 159 & 83.6 & 87.1 & +3.5 \\
    \rowcolor{gray!4}
     Tail-Instability-only
    & 40 & 60.0 & 77.5 & +17.5 \\
    \rowcolor{red!4}
    \textbf{Both failures}
    & 46 & 73.9 & 84.8 & +10.9 \\
    \bottomrule
    \end{tabular}
  \end{table}

Table~\ref{tab:diagnostic_effect} verifies the two failure modes diagnosed in Sec.~3.2. Relative to the SFT initialization, vanilla GRPO increases blank-image accuracy, same-answer rate, normalized trajectory similarity, and late-rise rate, indicating stronger spurious grounding and more severe tail drift. In contrast, ProSR consistently reduces these indicators, showing improved visual dependence and trajectory stability. Specifically, lower \(\mathrm{SAR}\) and \(\mathrm{NTS}\) imply weaker counterfactual invariance under blank-image perturbation, while lower \(\mathrm{LRR}@0.1\) reflects less late-stage entropy re-rise.  Notably, the reduction in \(A_{\mathrm{blank}}\) is desirable here, because it means the model becomes less able to answer correctly after visual evidence is removed, indicating stronger reliance on the image rather than language priors.

Figure~\ref{fig:diagnostic_overview} provides complementary trajectory-level evidence. ProSR enlarges the counterfactual trajectory gap over vanilla GRPO in Fig.~\ref{fig:diagnostic_overview}(a), indicating more distinct original/blank reasoning dynamics. In Fig.~\ref{fig:diagnostic_overview}(b), most samples lie above the diagonal, showing reduced per-sample NTS under ProSR. Fig.~\ref{fig:diagnostic_overview}(c) further confirms that ProSR suppresses the late-stage entropy rise of vanilla GRPO.  Overall, the trajectory-level plots suggest that vanilla GRPO makes the model more invariant to blank-image perturbation and more unstable at the tail, whereas ProSR moves both behaviors back toward the SFT regime.

  Table~\ref{tab:failure_breakdown} further shows that these gains are concentrated on failure-prone cases. We group samples into clean, spurious-grounding-only
  (\(\mathrm{SAR}=1\) and \(\mathrm{NTS}>0.4\)), tail-instability-only (\(\mathrm{LRR}@0.1>0\)), and both-failure cases. ProSR yields the largest improvements on
  the tail-instability-only and both-failure groups while changing clean samples only marginally, suggesting that it mainly repairs the cases where vanilla GRPO
  exhibits the strongest process degradation.

  \subsection{Ablation Study}
  \label{sec:ablation}

  We perform ablations to isolate the contribution of each process reward. Starting from the same SFT initialization, we compare vanilla GRPO, GRPO with only the Counterfactual
  Invariance Penalty, GRPO with only the Tail Drift Penalty, and ProSR.

\begin{table}[t]
  \centering
  \small
  \setlength{\tabcolsep}{5.2pt}
  \renewcommand{\arraystretch}{1.15}
  \caption{
  Ablation of the two process rewards.
  Higher is better for Avg.; lower is better for other.
  }
  \label{tab:ablation_main}
  \vspace{3pt}
  \begin{tabular}{lcccc}
  \toprule
  \rowcolor{gray!10}
  \textbf{Method} &
  \(\boldsymbol{\mathrm{Avg.}}\uparrow\) &
  \(\boldsymbol{\mathrm{SAR}}\downarrow\) &
  \(\boldsymbol{\mathrm{NTS}}\downarrow\) &
  \(\boldsymbol{\mathrm{LRR}@0.1}\downarrow\) \\
  \midrule
  Vanilla GRPO
  & 64.0 & 0.5792 & 0.4516 & 0.1792 \\
  \rowcolor{gray!4}
  + Counterfactual Invariance
  & 67.1 & 0.4375 & 0.3562 & 0.1084 \\
  \rowcolor{gray!4}
  + Tail Drift
  & 66.4 & 0.5228 & 0.4191 & 0.0526 \\
    \rowcolor{red!4}
  \textbf{ProSR}
  & \textbf{69.3} & \textbf{0.3379} & \textbf{0.2916} & \textbf{0.0017} \\
  \bottomrule
  \end{tabular}
\end{table}

  Table~\ref{tab:ablation_main} shows that the two shaping terms play complementary roles. Adding only the Counterfactual Invariance Penalty mainly improves the diagnostics
  related to visual dependence, including \(\mathrm{SAR}\) and \(\mathrm{NTS}\), whereas adding only the Tail Drift Penalty mainly reduces
  \(\mathrm{LRR}@0.1\). Combining both terms yields the best overall performance and the strongest process-level reliability, confirming that the two rewards target distinct yet
  complementary failure modes.

  \subsection{Qualitative Analysis}
  \label{sec:qualitative_analysis}

\begin{figure}[ht]
    \centering
    \includegraphics[width=\linewidth]{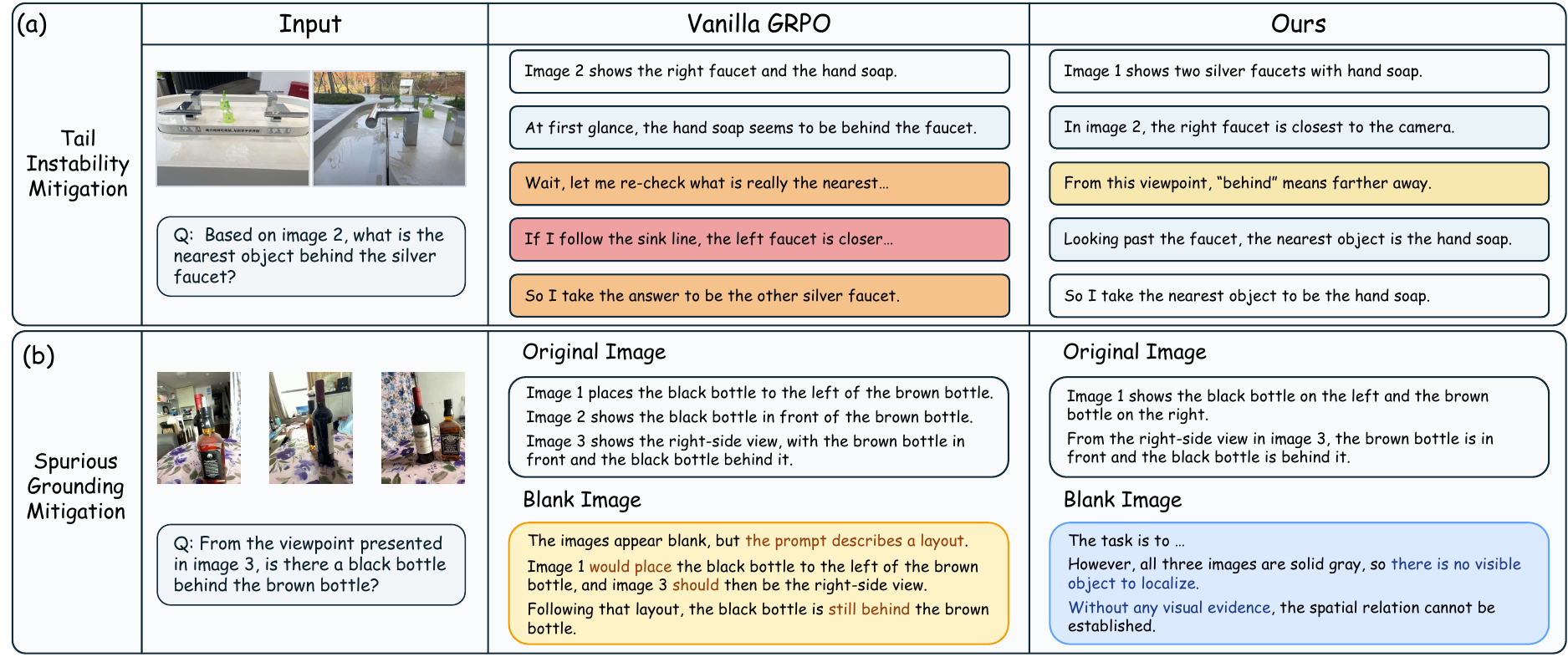}
    \caption{Qualitative comparison of CoT outputs. (a) A tail-instability case, where text color intensity indicates mean entropy. (b) A spurious-grounding case, where our method yields more visually grounded reasoning under the original image and less confident content under the blank-image
  probe.}
    \label{fig:qualitative_cases}
  \end{figure}

    We present two qualitative cases in Fig.~\ref{fig:qualitative_cases}, with token color indicating entropy intensity. In the spurious grounding case, vanilla GRPO
  keeps a similar answer and reasoning pattern even under the blank-image probe, whereas ProSR makes the trace more sensitive to missing visual evidence. In the
  tail drift case, vanilla GRPO shows late-stage entropy re-rise and unstable reasoning, while ProSR produces a smoother and more coherent trajectory. These
  examples match the quantitative diagnostics.

 \section{Conclusion}
  \label{sec:conclusion}

    We present ProSR, a diagnosis-driven framework for improving the process-level reliability of VLMs in spatial reasoning. By identifying
  Spurious Grounding and Tail Instability in vanilla GRPO and translating them into the Counterfactual Invariance Penalty and Tail Drift Penalty, ProSR improves
  both benchmark performance and reasoning-process quality. While our study is currently limited to one base model and entropy-based process signals, the results
  suggest that diagnosis-driven process shaping is a promising direction for more reliable visually grounded reasoning.

\clearpage
\bibliography{reference}

\clearpage
\appendix

 \section{Data Construction and Filtering}
  \label{app:data}

  \subsection{Source Benchmarks and Coverage}

  We construct the spatial CoT data by prompting Gemini-3.1-Pro-Preview to generate concise, visually grounded rationales for spatial reasoning questions drawn
  from \textsc{MindCube}, \textsc{SenseNova}, and \textsc{SPAR}. The original question text, associated image(s), and answer options are preserved, while the
  teacher model is only used to annotate reasoning traces. Table~\ref{tab:source_coverage} summarizes the source benchmarks and their main spatial coverage.

  \begin{table}[ht]
  \centering
  \small
  \setlength{\tabcolsep}{4.5pt}
  \renewcommand{\arraystretch}{1.12}
  \caption{Source benchmarks used for CoT construction.}
  \label{tab:source_coverage}
  \begin{tabular}{lccc}
  \toprule
  Source benchmark & Main spatial phenomena & Raw pool & Filtered subset \\
  \midrule
  MindCube & relative relation, viewpoint, metric reasoning & 9,469 & 4,850 \\
  SenseNova & relative relation, cross-view, generic scene queries & 9,813 & 4,068 \\
  SPAR & viewpoint, cross-view, metric/depth reasoning & 9,991 & 3,857 \\
  \bottomrule
  \end{tabular}
  \end{table}

  \subsection{Teacher Prompting Protocol}

  We do not ask the teacher model for long generic explanations. Instead, the prompt is designed to elicit short reasoning traces that explicitly identify the
  target spatial relation, rely only on task-relevant visual evidence, and end in a standardized answer format. Table~\ref{tab:prompt_principles} summarizes the
  main prompt design principles.

  \begin{table}[ht]
  \centering
  \small
  \setlength{\tabcolsep}{4.8pt}
  \renewcommand{\arraystretch}{1.12}
  \caption{Key design principles encoded in the teacher prompt.}
  \label{tab:prompt_principles}
  \begin{tabular}{lll}
  \toprule
  Component & Purpose & Expected behavior \\
  \midrule
  Task focus & Avoid generic captioning & Identify the target relation first \\
  Evidence grounding & Enforce visual support & Cite visible objects or views \\
  Direct spatial anchors & Strengthen relation reasoning & Use left/right/viewpoint/markers \\
  Conciseness & Prevent verbose loops & Keep 3--6 short steps \\
  Anti-fabrication & Suppress hallucination & Mention ambiguity only when needed \\
  Output format & Facilitate parsing & Output \texttt{<think>} and \texttt{<answer>} only \\
  \bottomrule
  \end{tabular}
  \end{table}

 \subsection{Quality Filtering Rules}

  After generation, we apply rule-based filtering to improve data quality. We retain only answer-correct samples, remove overlong or over-short reasoning traces,
  discard repeated self-revision, and require sufficient spatial grounding. Concretely, we define \(\mathrm{reconsider\_count}\) as the total number of occurrences
  of reconsideration markers such as ``wait'', ``let me reconsider'', ``let me re-examine'', ``let me re-think'', ``let me think again'', ``let me re-evaluate'',
  ``let me revisit'', ``on second thought'', ``let me check'', ``let me look at'', and ``hmm''. To quantify whether a rationale is spatially grounded, we compute a
  spatial-anchor ratio based on a predefined lexicon of explicit spatial expressions. Given a reasoning trace with token sequence \(\mathbf{w}=(w_1,\dots,w_T)\),
  we define
  \begin{equation}
  \rho_{\mathrm{anchor}}
  =
  \frac{1}{T}
  \sum_{t=1}^{T}
  \mathbf{1}[w_t \in \mathcal{V}_{\mathrm{sp}}],
  \end{equation}
  where \(\mathcal{V}_{\mathrm{sp}}\) denotes the spatial lexicon. Concretely, the spatial lexicon covers explicit relative relations (e.g., left/right, above/
  below, in front of/behind, beside), ordinal positions (e.g., leftmost/rightmost, topmost/bottommost), directional and viewpoint cues (e.g., north/south/east/
  west, clockwise/counterclockwise, viewpoint, perspective, camera, facing), and coordinate- or distance-like expressions (e.g., row/column indices, \((x,y)\)
  pairs, pixel-, meter-, or degree-based references). We count all matched patterns in the reasoning trace and normalize by word count to obtain
  \(\rho_{\mathrm{anchor}}\).

  \begin{table}[ht]
  \centering
  \small
  \setlength{\tabcolsep}{4.6pt}
  \renewcommand{\arraystretch}{1.12}
  \caption{Rule-based filtering criteria used for SFT data construction.}
  \label{tab:filter_rules}
  \begin{tabular}{lll}
  \toprule
  Signal & Criterion & Intuition \\
  \midrule
  Answer correctness & predicted answer matches reference & remove unfaithful reasoning traces \\
  CoT length & \(40 \le \text{length} \le 400\) & avoid trivial or excessively verbose traces \\
  Reconsider count & \(\mathrm{reconsider\_count} \le 2\) & suppress circular self-revision \\
  Repeated sentences & repeated\_sentence\_count \(\le 3\) & remove degenerate repetition \\
  Spatial grounding & \(\rho_{\mathrm{anchor}} \ge 0.04\) & retain spatially anchored reasoning \\
  \bottomrule
  \end{tabular}
  \end{table}

  \subsection{Filtering Effects}

  Table~\ref{tab:filtering_stats} summarizes the effect of filtering on the three source-specific CoT pools. The retained subsets exhibit stronger spatial-anchor
  density and generally lower reconsider frequency, indicating that the filtering stage removes weakly grounded or unstable reasoning traces before SFT.

  \begin{table}[ht]
  \centering
  \small
  \setlength{\tabcolsep}{4.2pt}
  \renewcommand{\arraystretch}{1.1}
  \caption{Filtering effects on source-specific CoT pools.}
  \label{tab:filtering_stats}
  \begin{tabular}{lcccc}
  \toprule
  Source & Keep rate & CoT length & Reconsider count & Spatial anchor ratio \\
  \midrule
  MindCube & 51.2\% & 204.9\(\rightarrow\)237.5 & 0.009\(\rightarrow\)0.006 & 0.113\(\rightarrow\)0.117 \\
  SenseNova & 41.5\% & 141.9\(\rightarrow\)139.9 & 0.005\(\rightarrow\)0.001 & 0.095\(\rightarrow\)0.108 \\
  SPAR & 38.6\% & 139.5\(\rightarrow\)136.6 & 0.003\(\rightarrow\)0.004 & 0.068\(\rightarrow\)0.073 \\
  \bottomrule
  \end{tabular}
  \end{table}

  \subsection{Final Dataset Composition}
  \label{app:task_comp}

  The final SFT initialization is trained on 22,135 filtered spatial CoT samples, and the reinforcement learning stage uses 44,500 spatial reasoning samples
  from the same source pool with additional expanded examples. The SFT set is therefore a filtered subset of the broader GRPO source pool. Table~\ref{tab:task_composition}
  summarizes the unified task-family composition of the final SFT set, the GRPO training pool, and the 480-example diagnostic subset. These unified families are
  intentionally coarse and merge several source-specific subtypes. In particular, \emph{Generic Spatial Grounding} includes categories such as
  viewpoint-conditioned queries, action-outcome questions, generic scene MCQs, and other source-specific spatial queries that are not cleanly expressed by the
  other four families. The diagnostic subset is additionally source-balanced, containing 160 examples each from \textsc{MindCube}, \textsc{SenseNova}, and
  \textsc{SPAR}. All benchmark evaluations are conducted on held-out test splits that are disjoint from both training stages.

  \begin{table}[ht]
  \centering
  \small
  \setlength{\tabcolsep}{4.5pt}
  \renewcommand{\arraystretch}{1.12}
  \caption{Unified task-family composition of the training and diagnostic pools.}
  \label{tab:task_composition}
  \begin{tabular}{lccc}
  \toprule
  Task family & SFT (22,135) & GRPO (44,500) & Diagnostic (480) \\
  \midrule
  Relative Spatial Relation & 8101 (36.6\%) & 13748 (30.9\%) & 45 (9.4\%) \\
  Viewpoint / Camera Transformation & 2686 (12.1\%) & 4768 (10.7\%) & 82 (17.1\%) \\
  Cross-view Correspondence / Matching & 1882 (8.5\%) & 4723 (10.6\%) & 73 (15.2\%) \\
  Metric / Distance / Depth & 2517 (11.4\%) & 4606 (10.4\%) & 84 (17.5\%) \\
  Generic Spatial Grounding & 6949 (31.4\%) & 16655 (37.4\%) & 196 (40.8\%) \\
  \bottomrule
  \end{tabular}
  \end{table}

  \subsection{Full System Prompt}

  The complete system prompt used for teacher annotation is shown below in a code-style box for readability.

\begin{center}
\setlength{\fboxsep}{8pt}
\fbox{%
\begin{minipage}{0.90\linewidth}
\raggedright
\footnotesize

You are an expert in spatial reasoning and visual understanding. Your goal is not to produce long explanations, but to generate concise, grounded, high-quality reasoning that can be used as training data.

\vspace{0.5em}
Follow these requirements:

\vspace{0.5em}
\textbf{1. State the task briefly first:}\\
Use one sentence to identify the spatial relation, movement direction, viewpoint correspondence, or target that must be determined.

\vspace{0.5em}
\textbf{2. Use only necessary evidence:}\\
Describe only the images and objects that are truly useful for solving the question. Do not summarize every image just for completeness.

\vspace{0.5em}
\textbf{3. Every step must be grounded:}\\
Each reasoning step must explicitly rely on visible objects, viewpoints, or relative spatial relations. Do not write vague summary sentences without concrete spatial support.

\vspace{0.5em}
\textbf{4. Prefer direct spatial anchors:}\\
Prefer direct relations such as left, right, above, below, in front of, behind, beside, clockwise, counterclockwise, and from image X's viewpoint. Introduce a coordinate system or global directions only when it is truly necessary for multi-view integration, rotation, or direction mapping. For tasks involving marked points, colored bounding boxes, depth, distance, coordinates, or cross-view matching, explicitly refer to the relevant markers and views. Do not invent hidden geometry or unseen object locations.

\vspace{0.5em}
\textbf{5. Keep the reasoning short and effective:}\\
Aim for 3 to 6 short steps. Do not repeat descriptions, do not keep changing your mind, and do not loop without new evidence.

\vspace{0.5em}
\textbf{6. Do not fabricate:}\\
If the visual evidence is insufficient, do not invent nonexistent objects, directions, or layouts. If there is slight ambiguity, mention it briefly and then answer based on the strongest available evidence.

\vspace{0.5em}
\textbf{7. End with a clear conclusion:}\\
The final answer must be exactly one letter: A, B, C, or D.

\vspace{0.7em}
The output format must be exactly:

\vspace{0.3em}
\begin{minipage}{0.95\linewidth}
\ttfamily
\textless{}think\textgreater{}\\
{[concise, grounded, step-by-step reasoning]}\\
\textless{}/think\textgreater{}\\
\textless{}answer\textgreater{}X\textless{}/answer\textgreater{}
\end{minipage}

\vspace{0.7em}
Where X must be exactly one of A, B, C, or D. Do not output anything after the \texttt{\textless{}answer\textgreater{}\textless{}/answer\textgreater{}} tag.

\end{minipage}%
}
\end{center}

  \section{Additional Experiments}
   \subsection{Effect of CoT Data Filtering}
  \label{sec:cot_filter_ablation}

   \begin{table}[ht]
    \centering
    \small
    \setlength{\tabcolsep}{5pt}
    \renewcommand{\arraystretch}{1.1}
    \caption{Effect of CoT data filtering for SFT initialization. Average is computed over the four benchmarks shown here.}
    \label{tab:cot_filter_ablation}
    \begin{tabular}{lccccc}
    \toprule
    Method & Avg.$\uparrow$ & MindCube & ViewSpatial & EmbSpatial & SPAR-Bench \\
    \midrule
    Unconstrained CoT SFT & 59.4 & 64.9 & 43.8 & 77.1 & 51.7 \\
    Filtered Spatial CoT SFT & 61.7 & 68.4 & 45.6 & 78.6 & 54.3 \\
    \bottomrule
    \end{tabular}
  \end{table}

   We further investigate whether the quality of the SFT initialization depends on the spatial grounding of the CoT supervision data. To isolate this factor, we
  compare our filtered spatial CoT set with a same-size unconstrained CoT variant generated from the same base model but without the spatial-anchor filtering
  rules. As shown in Table~\ref{tab:cot_filter_ablation}, all models are then initialized with SFT under the same training budget and evaluated on the same spatial
  benchmarks.

  The results indicate that filtering materially improves both downstream accuracy and process reliability. In particular, the unconstrained CoT variant tends to
  produce more verbose but less spatially anchored rationales, which weakens the initialization effect for subsequent GRPO. By contrast, our filtered data yields a
  more compact and visually grounded reasoning prior, leading to better optimization stability and stronger final performance. This suggests that the benefit of
  our framework does not come solely from using CoT supervision, but from constructing CoT data that is explicitly aligned with spatial reasoning and visual
  evidence.

  \subsection{Sensitivity to Reward Weights}
  \label{sec:weight_sensitivity}

  We study the sensitivity of ProSR to the weights of the two shaping terms. Starting from the same SFT initialization, we vary one reward weight at
  a time while keeping the other training settings fixed, and report both benchmark performance and process-level diagnostics.

   \begin{table}[ht]
    \centering
    \small
    \setlength{\tabcolsep}{5pt}
    \renewcommand{\arraystretch}{1.1}
    \caption{Sensitivity to the reward weights. Higher is better for Avg., while lower is better for the diagnostic metrics.}
    \label{tab:weight_sensitivity}
     \begin{tabular}{lcccc}
  \toprule
  Setting & Avg.$\uparrow$ & $\mathrm{SAR}\downarrow$ & $\mathrm{NTS}\downarrow$ & $\mathrm{LRR}@0.1\downarrow$ \\
  \midrule
  $\lambda_{\mathrm{cf}}=0.05$ & 68.5 & 0.3564 & 0.3058 & 0.0069 \\
  $\lambda_{\mathrm{cf}}=0.10$ & \textbf{69.3} & \textbf{0.3379} & \textbf{0.2916} & \textbf{0.0017} \\
  $\lambda_{\mathrm{cf}}=0.20$ & 69.0 & 0.3451 & 0.2974 & 0.0038 \\
  \midrule
  $\lambda_{\mathrm{drift}}=0.05$ & 68.8 & 0.3442 & 0.2968 & 0.0186 \\
  $\lambda_{\mathrm{drift}}=0.10$ & \textbf{69.3} & \textbf{0.3379} & \textbf{0.2916} & \textbf{0.0017} \\
  $\lambda_{\mathrm{drift}}=0.20$ & 69.1 & 0.3406 & 0.2939 & 0.0049 \\
  \bottomrule
  \end{tabular}
  \end{table}

    Table~\ref{tab:weight_sensitivity} shows that the proposed method is reasonably stable across a moderate range of reward weights. Increasing either shaping term
  from a weak setting improves both benchmark performance and process reliability, while overly strong penalties lead to slightly worse overall accuracy. The best
  trade-off is achieved around $\lambda_{\mathrm{cf}}=0.1$ and $\lambda_{\mathrm{drift}}=0.1$, which we adopt in all main experiments. These results suggest that
  the gains of our method are not tied to a narrow hyperparameter choice.

  \subsection{Sensitivity to Diagnostic Thresholds}
\begin{figure}[t]
    \centering
    \includegraphics[width=\linewidth]{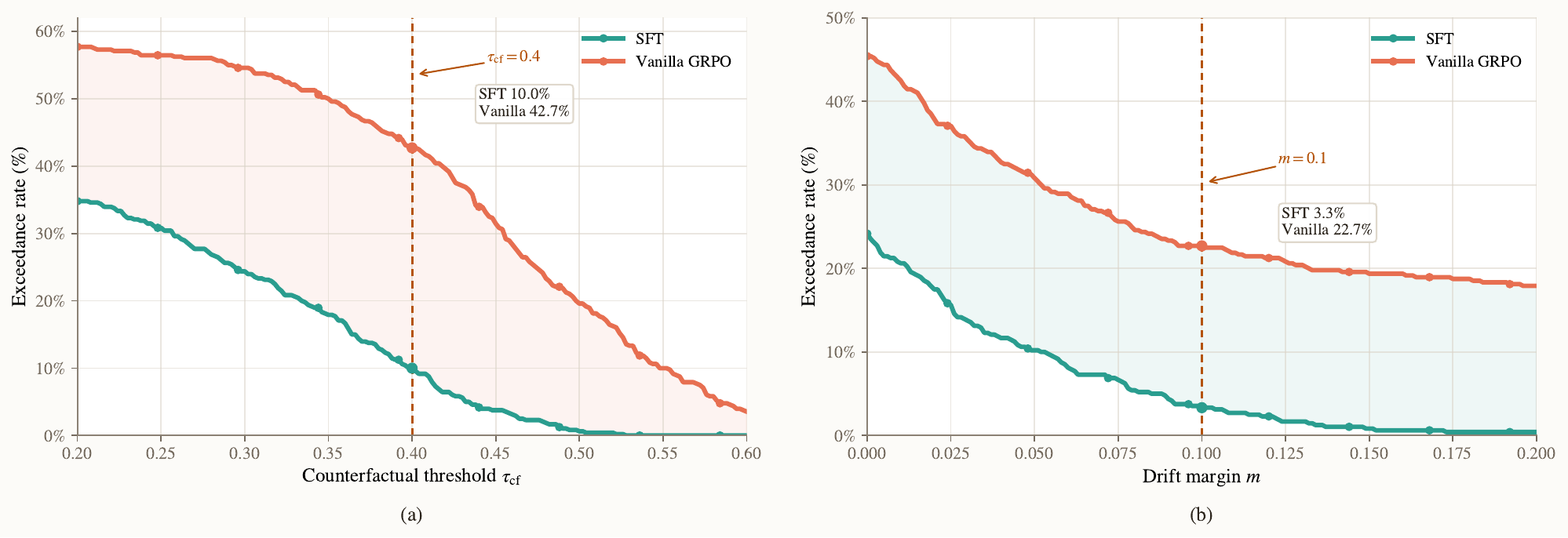}
    \caption{
    Sensitivity of the diagnostic thresholds used in the counterfactual and drift probes.
    (a) Exceedance rate under the counterfactual similarity threshold \(\tau_{\mathrm{cf}}\).
    (b) Exceedance rate under the late-rise margin \(m\).
    The separation between SFT and vanilla GRPO remains stable across a moderate range of cutoff values.
    }
    \label{fig:threshold_sensitivity}
  \end{figure}

    We further examine the robustness of the diagnostic thresholds used in our process-level analysis. Specifically, we scan the counterfactual similarity cutoff
  \(\tau_{\mathrm{cf}}\) and the late-rise margin \(m\) over a moderate range of values, and measure the fraction of diagnostic samples that exceed each threshold
  under SFT and vanilla GRPO. As shown in Fig.~\ref{fig:threshold_sensitivity}, the separation between the two models remains stable across a broad interval,
  indicating that our diagnosed failure modes are not artifacts of a particular cutoff choice. The selected defaults, \(\tau_{\mathrm{cf}}=0.4\) and \(m=0.1\), lie
  in a numerically stable region and provide a reasonable operating point for the main experiments.

\subsection{A Boundary Case in Egocentric Remapping}
 \begin{figure}[t]
    \centering
    \includegraphics[width=\linewidth]{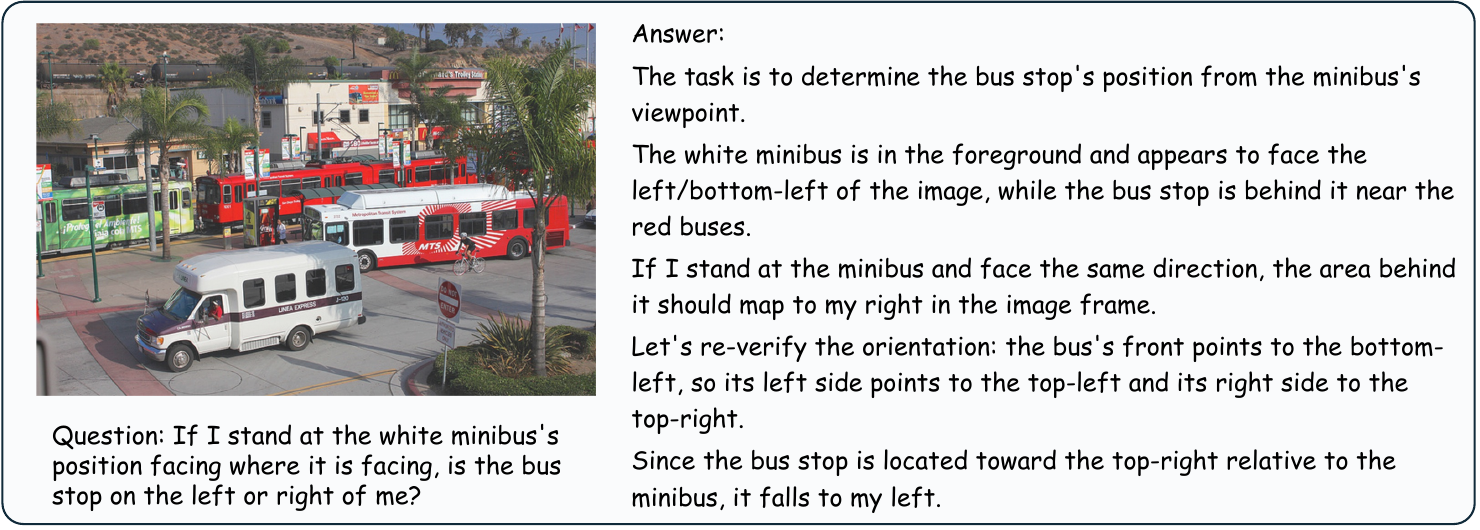}
    \caption{
    A representative failure case on \textsc{3DSRBench}. The model correctly identifies the minibus and the bus stop and explicitly re-checks its viewpoint mapping, but still
  inverts the final left/right relation after remapping to the observer's frame.
    }
    \label{fig:boundary_case}
  \end{figure}

  The failure case as shown in figure~\ref{fig:boundary_case} highlights the current boundary of ProSR. Although the model can correctly identify the relevant objects and even performs explicit self-rechecking, it may
  still fail on fine-grained egocentric remapping, where the answer depends on accurately transforming an object relation into the observer's local left/right frame. In the shown
  3DSRBench example, the model recovers the scene structure but still inverts the final lateral relation after re-anchoring the viewpoint, suggesting that process shaping improves
  grounding and stability but does not fully solve precise coordinate transformation under complex spatial layouts. Such cases indicate that the remaining challenge lies in
  robustly coupling visual evidence with viewpoint-aware geometric reasoning, especially when the correct answer depends on subtle perspective shifts rather than direct spatial
  cues.

\subsection{Computational Cost and Resources}
  \label{sec:computational_cost}

  Our method introduces a modest additional training cost. The counterfactual invariance term requires one extra blank-image rollout per prompt during training,
  which corresponds to approximately $12.5\%$ extra rollout-generation cost when $K=8$. In practice, this overhead is limited to the RL stage and does not require
  any additional annotations or auxiliary reward models. All experiments are run on 32 H20 GPUs; SFT takes about 1.5 hours per epoch (about 48 GPU-hours), while GRPO takes about 41 hours per epoch (about 1,312 GPU-hours).

 \section{Limitations}

  Our process-level diagnostics are based on entropy trajectories and blank-image counterfactual probing, so they should be interpreted as practical proxies rather
  than direct causal measurements of visual grounding. In particular, similar entropy trajectories do not necessarily imply identical internal reasoning
  mechanisms, and blank-image sensitivity is estimated from a single matched probe and therefore remains somewhat noisy. We thus treat these diagnostics as
  complementary evidence alongside benchmark accuracy and qualitative inspection, rather than as definitive proof of faithful visual reasoning.
  
  \section{Broader Impacts}

This work aims to improve the reliability of spatial reasoning in vision-language models by encouraging stronger visual grounding and more stable reasoning trajectories. It may benefit applications such as embodied agents, navigation, assistive systems, and visual instruction following, where accurate spatial understanding is important. However, more capable spatial reasoning models should still be deployed with caution in safety-critical or privacy-sensitive scenarios, as improved reasoning does not fully eliminate hallucination, bias, or misuse risks.


\end{document}